\documentclass[letterpaper]{article} 
\usepackage{aaai2027}  
\nocopyright

\usepackage[hyphens]{url}  
\usepackage{graphicx} 
\usepackage{natbib}  
\usepackage{caption} 
\usepackage{algorithm}
\usepackage{algorithmic}
\usepackage{amsmath}
\usepackage{amssymb}
\usepackage{newfloat}
\usepackage{listings}
\DeclareCaptionStyle{ruled}{labelfont=normalfont,labelsep=colon,strut=off} 
\floatstyle{ruled}
\newfloat{listing}{tb}{lst}{}
\floatname{listing}{Listing}

\usepackage{booktabs}

\title{DASH: Decoupled Adaptive Surrogate - Acquisition Harness\\ for Automated Bayesian Optimization}

\author{Changquan Zhao\textsuperscript{\rm 1}\equalcontrib,
        Yuxiang Sun\textsuperscript{\rm 2}\equalcontrib,
        Ruihao Zhu\textsuperscript{\rm 3}\corresponding,
        Cheng Hua\textsuperscript{\rm 2}\corresponding,
        Yulian He\textsuperscript{\rm 1}\corresponding}

\affiliations{
    \textsuperscript{\rm 1}
    Global College, Shanghai Jiao Tong University,
    Shanghai, China\\
    \textsuperscript{\rm 2}
    Antai College of Economics and Management,
    Shanghai Jiao Tong University,
    Shanghai, China\\
    \textsuperscript{\rm 3}
    SC Johnson College of Business,
    Cornell University,
    Ithaca, NY, USA\\
    chomaster@sjtu.edu.cn,
    scscsc04@sjtu.edu.cn,
    ruihao.zhu@cornell.edu\\
    cheng.hua@sjtu.edu.cn,
    yulian.he@sjtu.edu.cn
}

\begin{document}

\maketitle

\begin{abstract}
Bayesian optimization (BO) relies on a surrogate model and an acquisition function, yet the most suitable choices vary across tasks and optimization stages. Automated Bayesian optimization (AutoBO) addresses this variability by adapting BO components online. However, existing AutoBO methods either adapt one component, leaving the other mismatched and creating a bottleneck, or jointly select surrogate--acquisition pairs under a shared criterion, overlooking their distinct roles:  surrogate selection depends on predictive reliability, whereas acquisition adaptation should respond to campaign context.
In this paper, we propose \textbf{DASH}, a \textbf{D}ecoupled \textbf{A}daptive \textbf{S}urrogate--Acquisition \textbf{H}arness for large-language- model (LLM)-enhanced AutoBO. DASH selects surrogates by predictive reliability, uncertainty calibration, and ranking consistency; its two-stage acquisition controller periodically reallocates quotas across acquisition functions, builds a BO shortlist accordingly, and delegates final selection to an LLM. DASH also incorporates an integrated harness, consisting of knowledge-guided warm start and structured memory, to ground optimization in domain knowledge and accumulated feedback. Across four chemical optimization tasks, DASH outperforms the best AutoBO baseline by 12.51\% in trajectory-level Acceleration Factor and 5.00\% in endpoint Enhancement Factor. Results remain strong across LLM backbones, and ablations verify the complementary contributions of all components. Full-table and behavioral contamination checks find no detectable evidence that direct benchmark memorization or source-cell leakage explains these gains.
\end{abstract}


\section{Introduction}

Bayesian optimization (BO) is a sample-efficient method for optimizing costly black-box objectives, achieving strong performance in applications such as chemical reaction optimization~\cite{shields2021bayesian} and hyperparameter tuning~\cite{snoek2012practical}. Traditional BO typically relies on a fixed surrogate model to fit the unknown response surface and a fixed acquisition strategy to select subsequent evaluations. However, it has been widely demonstrated that the optimal surrogate and acquisition strategy differ across problem instances  and evolve across optimization stages, making a static configuration suboptimal~\cite{wang2023recent}.

To address this, automated Bayesian optimization (AutoBO) has been proposed to adapt BO configurations during the optimization process. Early developments mainly focus on dynamically adjusting the acquisition strategy based on historical observations ~\cite{hoffman2011portfolio,vasconcelos2019past}. Although effective, they often overlook broader optimization context such as search progress, remaining budget, and domain knowledge. In addition, they leave the surrogate model fixed, which can still lead to model misspecification and degrade performance. Recently, with the help of large language models (LLMs), pilot studies have used LLMs to enhance AutoBO through kernel evolution~\cite{suwandi2025cake}, acquisition function switching~\cite{ngo2026adaptive}, and candidate or sampling guidance~\cite{liu2024large,yang2025reasoning}, achieving promising results. However, these methods still adapt only one component at a time. Unfortunately, a misaligned combination can still hurt the performance of the entire AutoBO loop.

As such, recent work studies their joint selection by retrospectively evaluating surrogate--acquisition pairs on the available data~\cite{park2025boost}. However, the two components play different roles in BO. The surrogate model estimates the response surface and uncertainty, whereas the acquisition strategy determines how to act under the current belief, budget, and search trajectory. Thus, surrogate adaptation is naturally data-driven, while acquisition adaptation is better treated as a context-driven decision. Jointly selecting surrogate--acquisition pairs under a single retrospective criterion tends to overlook their distinct roles and overfit the limited observations. 

Motivated by this observation, we propose \textbf{DASH}, a \textbf{D}ecoupled \textbf{A}daptive \textbf{S}urrogate-acquisition \textbf{H}arnessing framework for LLM-enhanced AutoBO. DASH selects the surrogate model by evaluating how well candidate models fit historical observations in both predictive mean and uncertainty. Meanwhile, it also leverages an LLM to adaptively allocate candidate quotas across acquisition functions and select the next evaluation from the resulting candidate pool. To support reliable LLM-guided acquisition and the broader AutoBO process, DASH incorporates essential harnessing mechanisms, including knowledge-guided warm start and structured memory system. The warm start incorporates domain knowledge into initialization, while the
memory records the optimization trajectory and reusable knowledge to ground
subsequent decisions in accumulated evidence.
Together, they provide informed initialization and persistent feedback,
improving both LLM decision-making and the reliability of the overall AutoBO
loop. Our main contributions can be summarized as follows.
\begin{itemize}
    \item We propose a novel decoupled AutoBO framework that
     adapts surrogate model and acquisition
    strategy through data-driven and LLM-driven paradigms,
    respectively.

    \item We develop DASH, which complements its decoupled
    controllers with essential harnessing mechanisms to support reliable LLM-enhanced AutoBO.

    \item We demonstrate consistent improvements over AutoBO and LLM-enhanced BO baselines across chemical reaction optimization and hyperparameter tuning.
\end{itemize}

\section{Problem Setup and Preliminaries}

We consider sequential black-box optimization over a mixed search space $\mathcal{X}$. Each candidate $x \in \mathcal{X}$ may contain continuous and categorical variables. At iteration $t$, the optimizer selects a candidate $x_t$, observes a scalar response $y_t = f(x_t) + \epsilon_t$, and updates the observation set $\mathcal{D}_t$. The goal is to identify a high-performing candidate under a limited evaluation budget.

BO addresses this problem by maintaining a probabilistic surrogate model over the unknown objective and using an acquisition function to decide where to evaluate next. Many classical BO methods use Gaussian processes (GPs) as surrogate models, since GPs provide both predictive means and uncertainty estimates. The surrogate model estimates the response surface and its uncertainty, while the acquisition function translates these estimates into a sampling decision that balances exploitation and exploration. Common acquisition functions include Expected Improvement (EI)~\cite{jones1998efficient}, Probability of Improvement (PI)~\cite{kushner1964new}, Upper Confidence Bound (UCB)~\cite{srinivas2010gaussian}, and Thompson Sampling (TS)~\cite{thompson1933likelihood}. When multiple evaluations can be
performed in parallel, EI and UCB can be extended to $q$ Expected
Improvement (qEI)~\cite{wang2020parallel} and $q$ Upper Confidence Bound
(qUCB)~\cite{wilson2018maximizing}, respectively. In standard BO, both the surrogate model and the acquisition function are often fixed before optimization begins.

In practical experimental spaces, however, the most suitable modeling and sampling strategies may change across tasks and optimization stages. Early iterations may require broader exploration and robust uncertainty estimates, whereas later iterations may benefit from more exploitative or locally refined search. This motivates an AutoBO setting in which BO components are adapted during the campaign. In this work, we distinguish two adaptation problems: surrogate-side adaptation, which determines which model is currently reliable for representing the observed data, and acquisition-side strategy adaptation, which determines how the optimizer should act under the current search context.

\section{Related Work}

\noindent\textbf{Traditional AutoBO.}
Traditional AutoBO methods typically adapt BO configurations using numerical
feedback from the optimization trajectory.
Early research primarily focused on adapting the acquisition function.
GP-Hedge~\cite{hoffman2011portfolio} constructs an acquisition function portfolio and updates the
selection probabilities through a Hedge-based online
mechanism, while No-PASt-BO~\cite{vasconcelos2019past} improves this strategy
by discounting earlier evaluations and normalizing portfolio
rewards. HEBO~\cite{cowen2022hebo} jointly optimizes an ensemble of acquisition functions and selects  the resulting Pareto-efficient candidate.
Other works explore dynamic schedules for switching between EI and
PI~\cite{benjamins2022pi}, as well as heuristic acquisition function
adaptation toward AutoBO~\cite{garrido2021automatic}.
More recently, BOOST~\cite{park2025boost} extends automatic configuration beyond the acquisition
function by retrospectively evaluating candidate kernel--acquisition pairs
on the available observations and selecting the pair expected to perform best
in subsequent optimization.

\noindent\textbf{LLM-Enhanced BO.}
In recent years, LLMs have been integrated into BO to leverage their 
world knowledge and reasoning capabilities for
optimization. One line of work uses LLMs to enhance individual BO components.
GOLLuM~\cite{rankovic2025gollum} combines LLM representations with GP-based optimization through
LLM-based deep kernels, CAKE~\cite{suwandi2025cake} uses LLMs to generate
and evolve GP kernels during optimization, and LMABO~\cite{ngo2026adaptive}
formulates acquisition function selection as an in-context decision-making
problem. Another line of work uses LLMs as broader assistants throughout the BO
process. LLAMBO~\cite{liu2024large} formulates BO as a natural language context for warm start,
surrogate modeling, and candidate sampling. Reasoning BO~\cite{yang2025reasoning} and BORA~\cite{cisse2025bora} further use LLMs for  hypothesis generation, sampling guidance, and real-time interpretation of the optimization process.

\section{Proposed DASH Framework}
\begin{figure*}[t]
    \centering
    \includegraphics[width=0.95\textwidth]{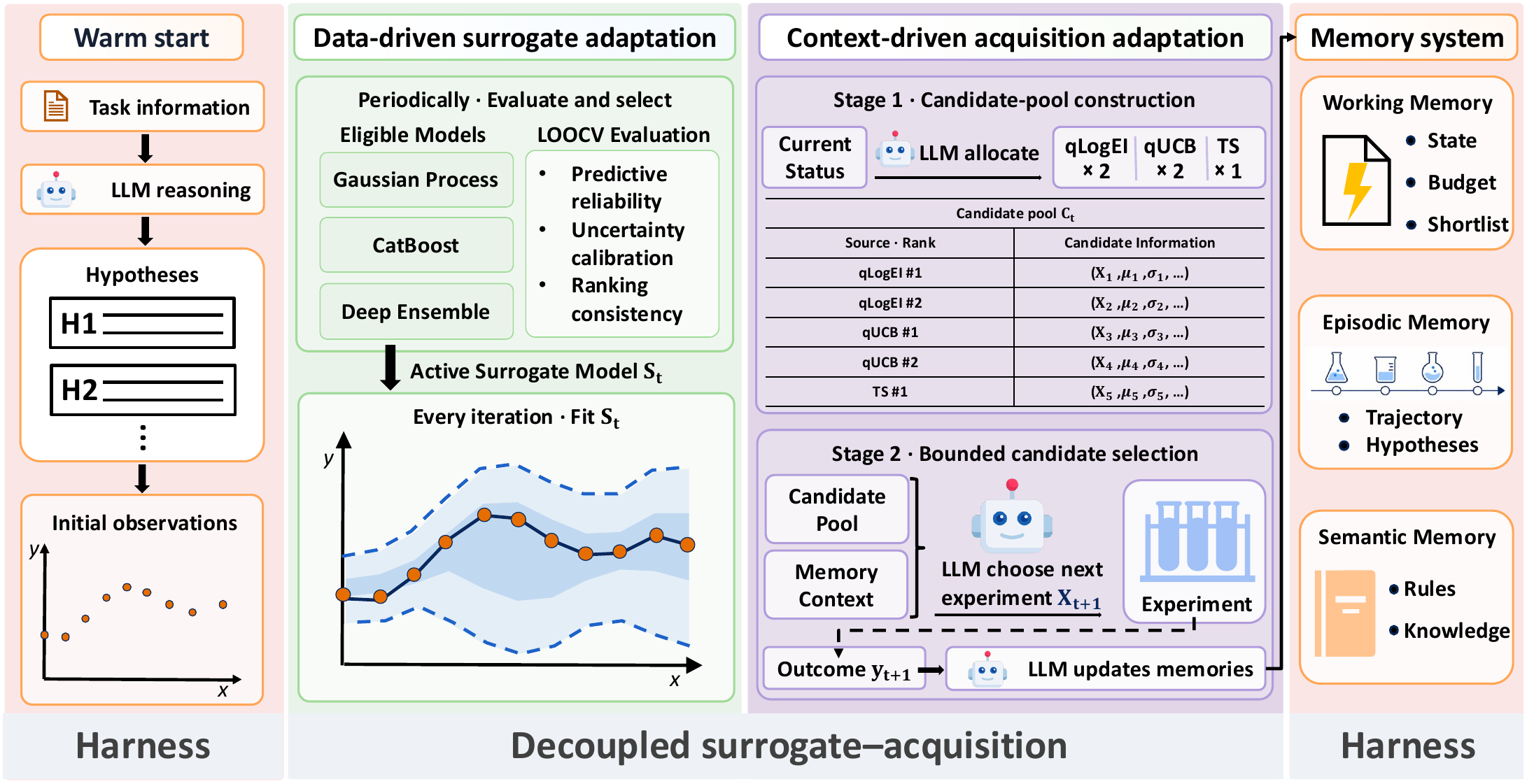}
    \caption{Overview of the proposed DASH framework.}
    \label{fig:workflow}
\end{figure*}

The overall framework of DASH is illustrated in Figure~\ref{fig:workflow}.
Given a specific optimization problem, DASH first draws on the LLM's world knowledge to formulate relevant hypotheses and stores them in structured memory. Guided by these hypotheses, it designs a warm start that balances the exploration and exploitation, yielding the initial observation set. After initialization, each BO iteration follows two decoupled adaptation paths. The data-driven path periodically evaluates candidate surrogate models using the accumulated observations and updates the active model when warranted. The context-driven path first uses the LLM to adapt the acquisition function configuration to the current optimization state and construct a candidate pool, and then asks the LLM to select the next evaluation point based on memory and
candidate-specific information. Once the result is observed, DASH reflects on the outcome and updates its memory by revising the hypotheses and distilling
reusable knowledge for subsequent decisions.

\subsection{Data-Driven Surrogate Model Adaptation}
The suitability of a surrogate model depends on both the optimization problem and the observations accumulated at the current stage. A reliable surrogate model should represent the response surface through an mean estimate and credible uncertainty quantification. The former determines which regions appear promising, whereas the latter indicates how strongly those predictions are supported and guides exploration of insufficiently observed regions. These properties can be assessed directly from how well a model predicts existing observations when they are held out. DASH therefore uses the collected data as empirical evidence for surrogate adaptation, periodically comparing the predictive performance of available models and updating the active surrogate  when the evidence supports a change.

To accommodate response surfaces with different structures, DASH maintains a diverse surrogate candidate pool, including mixed-variable Gaussian process (GP) models with different continuous and categorical kernels~\cite{williams2006gaussian}, CatBoost~\cite{prokhorenkova2018catboost}, and deep ensembles~\cite{lakshminarayanan2017simple}.

To balance timely improvements in predictive fit against unstable model switching, DASH evaluates the surrogate pool every $k$ iterations rather than at every iteration. At evaluation iteration $t$, let $\mathcal{D}_t=\{(x_j,y_j)\}_{j=1}^{n_t}$ denote the accumulated observations and $\mathcal{S}_t$ the set of eligible surrogates. For each $s_i\in\mathcal{S}_t$, DASH performs leave-one-out cross-validation (LOOCV). Specifically, it successively removes each observation $(x_j,y_j)$, fits $s_i$ on the remaining observations, and predicts the held-out input $x_j$ to obtain the predictive mean $\mu_{ij}$ and standard deviation $\sigma_{ij}$. These held-out predictions are then aggregated into three complementary surrogate-level evaluation signals.

\noindent\textbf{Predictive reliability.}
$f_{\mathrm{seq}}$ averages the predictive log likelihood across all held-out observations:
\begin{equation}
\begin{aligned}
&f_{\mathrm{seq}}(s_i)
=\frac{1}{n_t}\sum_{j=1}^{n_t}\ell_{ij},\\
&\ell_{ij}
=-\frac{1}{2}\log(2\pi\sigma_{ij}^{2})
-\frac{1}{2}
\left(\frac{y_j-\mu_{ij}}{\sigma_{ij}}\right)^2.
\end{aligned}
\end{equation}
The standardized residual term penalizes inaccurate predictive means, while the log variance term prevents the model from masking prediction
errors with excessively diffuse uncertainty. A higher $f_{\mathrm{seq}}$ therefore indicates better out-of-sample probabilistic predictive performance under LOOCV.

\noindent\textbf{Uncertainty calibration.} $f_{\mathrm{cal}}$ compares the empirical
coverage of the held-out predictive intervals with a target
coverage level $\rho$:
\begin{equation}
\begin{aligned}
f_{\mathrm{cal}}(s_i)
&=-\left|\operatorname{cov}(s_i)-\rho\right|,\\
\operatorname{cov}(s_i)
&=\frac{1}{n_t}\sum_{j=1}^{n_t}
\mathbf{1}\!\left[
y_j\in
\left[\mu_{ij}-z_{\rho}\sigma_{ij},
      \mu_{ij}+z_{\rho}\sigma_{ij}\right]
\right],
\end{aligned}
\end{equation}
where $z_\rho$ denotes the corresponding quantile of the standard normal distribution. A value closer to zero indicates better calibration, whereas a larger deviation suggests that the surrogate is either overconfident or overly conservative.

\noindent\textbf{Ranking consistency.} $f_{\mathrm{rank}}$ evaluates whether the
held-out predictive means preserve the ordering of the high-performing observations. Let $\mathcal{I}_t$ denote the indices of up to the top $K$ observations according to the optimization direction at iteration $t$. The score is defined as
\begin{equation}
f_{\mathrm{rank}}(s_i)
=\operatorname{Spearman}
\left(
\{\mu_{ij}\}_{j\in\mathcal{I}_t},
\{y_j\}_{j\in\mathcal{I}_t}
\right).
\end{equation}
A higher $f_{\mathrm{rank}}$ indicates that the surrogate more reliably distinguishes among the promising regions.

After computing the three surrogate-level metrics, DASH standardizes each metric across the eligible surrogate pool and combines them into the overall score
\begin{equation}
\begin{aligned}
f(s_i)={}&w_{\mathrm{seq}}z_{\mathrm{seq}}(s_i)
+w_{\mathrm{cal}}z_{\mathrm{cal}}(s_i)
+w_{\mathrm{rank}}z_{\mathrm{rank}}(s_i),
\end{aligned}
\end{equation}
where $z_q(s_i)$ denotes the standardized value of
$f_q(s_i)$ among the eligible surrogates, and
$w_{\mathrm{q}}$ determine the relative contributions of the three evaluation dimensions, \(q\in\{\mathrm{seq},\mathrm{cal},\mathrm{rank}\}\). Let $s_t^\star=\arg\max_{s_i\in\mathcal{S}_t}f(s_i)$ denote the highest scoring eligible surrogate and $s_t^{\mathrm{act}}$ the active surrogate. DASH switches to $s_t^\star$ only if the active surrogate fails the evaluation or $f(s_t^\star)-f(s_t^{\mathrm{act}})>\delta$; otherwise, it retains the active surrogate to avoid unstable switching.

\subsection{Context-Driven Acquisition Strategy Adaptation}
Conventional BO typically uses a fixed acquisition function and selects its top-ranked candidate at each iteration. This design has two limitations. First, a fixed acquisition function maintains the same exploration--exploitation preference as the search evolves, although the appropriate preference may change with the optimization progress. Second, the top-1 decision follows a single numerical ranking and cannot account for broader context, including recent outcomes, active hypotheses, domain knowledge, and accumulated experience. DASH therefore adopts a two-stage acquisition strategy. Stage~1 is triggered periodically to reconfigure the acquisition weights from the current optimization progress, after which the latest configuration is used to construct a candidate pool. Stage~2 then uses the LLM to select the next experiment from this pool by combining campaign context with candidate-specific evidence.

\noindent\textbf{Stage 1: Candidate-pool construction.}
DASH considers three standard acquisition functions. qLogEI favours refinement around regions that have already shown promise, qUCB supports controlled exploration when uncertainty remains informative, and TS encourages broader exploration when greater search diversity is needed.

At every BO iteration, DASH constructs a pool $\mathcal{C}_t$ containing $N$ unique candidates. Once a reconfiguration is triggered, the LLM receives a summary of the current optimization progress and determines how the $N$ candidate quotas are distributed across the three acquisition functions. The resulting allocation is used until the next reconfiguration. Each acquisition function contributes its highest-ranked candidates up to the assigned quota. Duplicate proposals are merged, and additional candidates are drawn from the subsequent acquisition rankings until the pool contains $N$ unique candidates.

For each retained candidate, DASH records its predictive mean and uncertainty together with a compact acquisition profile comprising its proposing functions, within-function ranks, and cross-function consensus. This information is passed to Stage~2 to support the bounded LLM selection.

\noindent\textbf{Stage 2: Bounded LLM candidate selection.}
Given $\mathcal{C}_t$, the LLM considers two complementary sources of evidence before selecting the next experiment. Campaign-level evidence is retrieved from memory and summarizes the optimization progress, active hypotheses, and knowledge accumulated during the search. Candidate-level evidence links each candidate to the surrogate belief through its predictive mean and uncertainty, and to the acquisition information through its sources, ranks and consensus. The LLM combines these signals with its pretrained scientific knowledge to assess each candidate's practical plausibility, predicted potential, and value to the current search trajectory. It then selects the most valuable candidate for the current stage as the next experiment.

\subsection{Harnessing Mechanisms}
To support reliable and effective LLM-guided AutoBO, DASH incorporates essential harnessing mechanisms including knowledge-guided warm start and structured memory system, strengthening both LLM decision quality and overall AutoBO performance.

\noindent\textbf{Knowledge-guided warm start.}
Given a task specification, DASH uses the LLM's world knowledge and reasoning capabilities to generate relevant hypotheses that identify promising search directions. Guided by these hypotheses, the LLM selects \(M\) initial points from the legal candidate space, balancing well-supported regions with plausible but uncertain alternatives while avoiding duplicate or concentrated choices. Their evaluations form the initial observation set for subsequent BO  iteration.

\noindent\textbf{Structured memory.}
Structured memory maintains optimization information at three levels: working memory represents the current optimization state, episodic memory records the optimization trajectory and the evolving status of active hypotheses, and semantic memory stores reusable knowledge derived during optimization. After each observation, DASH interprets the outcome, updates the current state and relevant hypothesis statuses, and consolidates the new reusable knowledge. This evolving context informs acquisition configuration and guides bounded selection from later BO shortlists. Together, warm start and structured memory ground LLM decisions in task knowledge and accumulated evidence while maintaining continuity across AutoBO.

\section{Experiments}
\subsection{Experimental Setup}

\noindent\textbf{Datasets.}
Following prior work, we evaluate the effectiveness and generalizability of DASH on eight benchmark tasks spanning two application domains: chemical optimization and hyperparameter optimization (HPO). The chemical suite comprises four representative datasets: 1) oxidative coupling of methane (OCM)~\cite{nguyen2020ocm}, a mixed-space task that jointly optimizes categorical catalyst compositions and continuous process conditions; 2) direct arylation reaction optimization (DAR)~\cite{shields2021bayesian}, which combines categorical chemical choices with continuous operating conditions; 3) oxygen evolution reaction electrocatalyst optimization (OER)~\cite{rohr2020benchmarking}, which searches a discretized continuous simplex of electrocatalyst compositions; and 4) Suzuki--Miyaura cross-coupling optimization (Suzuki)~\cite{perera2018platform}, which involves a categorical search space of high dimension and validity constraints defined by the dataset. The HPO suite comprises four tasks from HPOBench~\cite{eggensperger2021hpobench}: SVM, random forest, XGBoost, and neural network. These tasks cover continuous and integer hyperparameters.

\noindent\textbf{Baseline methods.}
We compare DASH with seven representative baselines covering the two major families: traditional AutoBO and LLM-enhanced BO methods.
Specifically, the traditional AutoBO baselines include:
1) GP-Hedge~\cite{hoffman2011portfolio}, which maintains an
acquisition function portfolio;
2) HEBO~\cite{cowen2022hebo}, which aggregates multiple
acquisition functions to select the next candidate; and
3) BOOST~\cite{park2025boost}, which retrospectively
evaluates and jointly selects surrogate-acquisition pairs.
The LLM-enhanced BO baselines include:
1) CAKE~\cite{suwandi2025cake}, which uses an LLM to generate
and evolve GP kernels;
2) LMABO~\cite{ngo2026adaptive}, which formulates
acquisition function selection as an in-context LLM decision;
3) Reasoning BO~\cite{yang2025reasoning}, which uses iterative LLM reasoning to guide candidate sampling; and
4) BORA~\cite{cisse2025bora}, which employs LLM  to identify promising regions.

\noindent\textbf{Evaluation metrics.}
To characterize the optimization process, we report best-so-far trajectories, which track the best objective value identified as the evaluation budget increases. Following prior work on benchmarking sequential experimental optimization~\cite{liang2021benchmarking}, we further use the Enhancement Factor (EF) and Acceleration Factor (AF) to summarize attained solution quality and overall optimization efficiency, respectively:
\[
\mathrm{EF}
=
\frac{y_{\mathrm{best}}-y_{\min}}
     {y_{\mathrm{global}}-y_{\min}},
\qquad
\mathrm{AF}
=
\frac{\mathrm{AUC}}
     {\mathrm{AUC}_{\mathrm{RS}}}.
\]
Here, $y_{\mathrm{best}}$ is the best objective value found within the evaluation budget, $y_{\mathrm{global}}$ and $y_{\min}$ denote the global optimum and minimum objective value in the corresponding benchmark dataset, respectively. $\mathrm{AUC}$ is the area under the best-so-far trajectory, and $\mathrm{AUC}_{\mathrm{RS}}$ is the mean AUC obtained by Random Search under the same budget. Higher EF indicates better final solution quality, while higher AF indicates faster optimization progress relative to random exploration.

\noindent\textbf{Implementation Details.}
All methods were evaluated over five independent seeds with a budget of 40 evaluations, comprising 10 initial observations and 30 BO iterations. Dataset evaluations were performed by table lookup, and all LLM-based methods used DeepSeek-V4-Pro with a temperature of 0.3 and reasoning enabled. Complete DASH configurations are reported in Supplementary Material.

\subsection{Main Results}

\begin{table*}[t]
\centering
\tiny
\setlength{\tabcolsep}{2pt}
\renewcommand{\arraystretch}{1.15}
\caption{AF and EF results. Values are mean$_{\pm \mathrm{std}}$ across dataset-level values. Best and second-best values in each row are bolded and underlined, respectively. Improvement is computed relative to the best non-DASH baseline in each row.}
\label{tab:af_ef_results}
\begin{tabular*}{\textwidth}{@{\extracolsep{\fill}}llccccccc|cc@{}}
\hline
Metric & Dataset & GP-Hedge & HEBO & BOOST & CAKE & LMABO & BORA & ReasoningBO & Ours & Improv. \\
\hline
AF & Chem Avg. & 1.1418$_{\pm 0.2112}$ & 1.1077$_{\pm 0.0929}$ & \underline{1.1716}$_{\pm 0.1263}$ & 1.0894$_{\pm 0.0824}$ & 1.0970$_{\pm 0.0724}$ & 1.0757$_{\pm 0.0922}$ & 1.1155$_{\pm 0.1184}$ & \textbf{1.3182}$_{\pm 0.3816}$ & +12.51\% \\
 & HPO Avg. & 0.9979$_{\pm 0.0121}$ & 1.0088$_{\pm 0.0098}$ & 1.0085$_{\pm 0.0126}$ & 1.0096$_{\pm 0.0045}$ & \underline{1.0105}$_{\pm 0.0033}$ & 1.0099$_{\pm 0.0124}$ & 1.0052$_{\pm 0.0126}$ & \textbf{1.0160}$_{\pm 0.0053}$ & +0.54\% \\
\hline
EF & Chem Avg. & 0.8883$_{\pm 0.1067}$ & 0.8434$_{\pm 0.2331}$ & \underline{0.9083}$_{\pm 0.1507}$ & 0.8614$_{\pm 0.1707}$ & 0.8289$_{\pm 0.2000}$ & 0.8514$_{\pm 0.2221}$ & 0.8746$_{\pm 0.1741}$ & \textbf{0.9538}$_{\pm 0.0695}$ & +5.00\% \\
 & HPO Avg. & 0.9939$_{\pm 0.0047}$ & 0.9917$_{\pm 0.0102}$ & 0.9923$_{\pm 0.0104}$ & 0.9955$_{\pm 0.0050}$ & 0.9930$_{\pm 0.0071}$ & \underline{0.9966}$_{\pm 0.0029}$ & 0.9901$_{\pm 0.0130}$ & \textbf{0.9975}$_{\pm 0.0018}$ & +0.09\% \\
\hline
\end{tabular*}
\end{table*}

\begin{figure*}[t]
    \centering
    \includegraphics[width=0.95\textwidth]{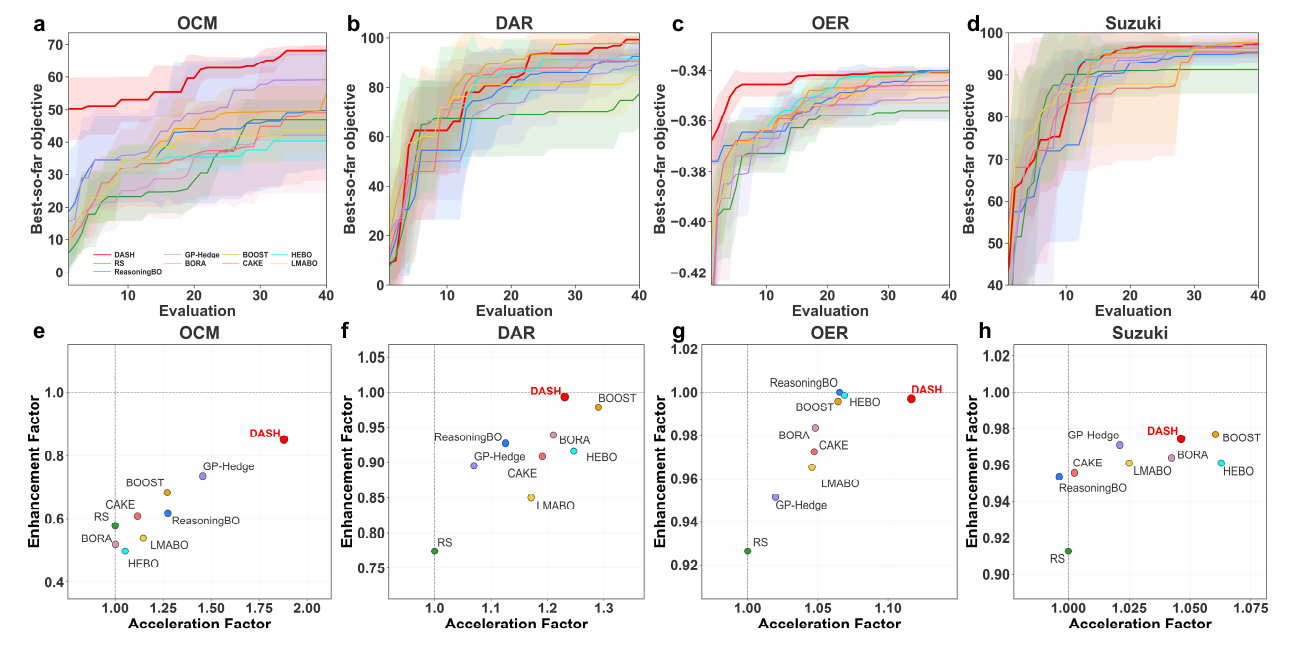}
    \caption{Results on the four chemical optimization benchmarks:
OCM, DAR, OER, and Suzuki.
(a--d) Mean best-so-far objective trajectories over 40
evaluations for DASH and the baseline methods, with shaded
regions indicating one standard deviation across runs.
(e--h) Joint comparison of AF and EF, Higher values on both axes indicate better optimization performance.}
    \label{fig:chemical_results}
\end{figure*}

\noindent\textbf{Overall performance.}
Table~\ref{tab:af_ef_results} summarizes the cross-task AF and EF results, while Figure~\ref{fig:chemical_results} presents the task-level optimization trajectories and the joint AF--EF comparison on the four chemical benchmarks. On the chemical optimization tasks, DASH achieves the highest average AF of $1.3182 \pm 0.3816$ and the highest average EF of $0.9538 \pm 0.0695$. Compared with BOOST, which selects the surrogate kernel and acquisition function as a coupled pair, these results correspond to relative improvements of $12.51\%$ in AF and $5.00\%$ in EF. The larger improvement in AF suggests that DASH provides its main advantage by improving the overall optimization trajectory, while also identifying final solutions of higher quality within the evaluation budget.

On the HPO benchmarks, DASH also obtains the highest average AF and EF, although the improvements over the strongest baselines are limited to $0.54\%$ and $0.09\%$, respectively.This smaller separation may reflect the limited discriminability of these HPO benchmarks, where the presence of many configurations with strong performance allows most methods to reach regions close to the optimum within the evaluation budget. We therefore focus the following analyses on the chemical optimization tasks, where the methods exhibit clearer performance differences. Detailed task-level results for the four HPO benchmarks are provided in Supplementary Material.

\noindent\textbf{Task-level behavior.}
The advantage of DASH is most pronounced on OCM, where it substantially outperforms all baselines in both AF and EF. On DAR, several methods progress rapidly during initialization, whereas DASH maintains stronger subsequent refinement and reaches the best final result. On OER, DASH achieves a substantially higher AF while retaining an EF close to those of the best-performing methods. On Suzuki, the differences among leading methods are smaller because most methods reach high-performing regions early and subsequently saturate, while DASH remains competitive in both AF and EF.

\subsection{Ablation Study}
\begin{table*}[t]
\centering
\scriptsize
\setlength{\tabcolsep}{3pt}
\renewcommand{\arraystretch}{1.18}
\caption{Ablation results on chemistry datasets. Values are mean$_{\pm \mathrm{std}}$ over runs for individual datasets; Average reports mean$_{\pm \mathrm{std}}$ over dataset-level means. $\Delta$ vs. Ours reports relative change in the Average column.}
\label{tab:ablation}
\begin{tabular*}{\textwidth}{@{\extracolsep{\fill}}llcccc|cc@{}}
\hline
Metric & Method & OCM & DAR & OER & Suzuki & Average & $\Delta$ vs. Ours \\
\hline
AF & Ours & 1.8791$_{\pm 0.1154}$ & 1.2306$_{\pm 0.1017}$ & 1.1166$_{\pm 0.0123}$ & 1.0463$_{\pm 0.0215}$ & 1.3182$_{\pm 0.3816}$ & -- \\
 & w.o. Surr. Switch & 1.7447$_{\pm 0.2494}$ & 1.2563$_{\pm 0.0556}$ & 1.1097$_{\pm 0.0182}$ & 1.0154$_{\pm 0.0288}$ & 1.2815$_{\pm 0.3243}$ & -2.78\% \\
 & w.o. LLM Acq. & 1.8498$_{\pm 0.2207}$ & 1.2201$_{\pm 0.0717}$ & 1.0867$_{\pm 0.0295}$ & 1.0151$_{\pm 0.0535}$ & 1.2929$_{\pm 0.3808}$ & -1.91\% \\
 & w.o. Harness & 1.0280$_{\pm 0.2284}$ & 1.0522$_{\pm 0.2273}$ & 1.0649$_{\pm 0.0238}$ & 0.9514$_{\pm 0.0674}$ & 1.0241$_{\pm 0.0509}$ & -22.31\% \\
\hline
EF & Ours & 0.8505$_{\pm 0.0186}$ & 0.9932$_{\pm 0.0150}$ & 0.9969$_{\pm 0.0037}$ & 0.9744$_{\pm 0.0112}$ & 0.9538$_{\pm 0.0695}$ & -- \\
 & w.o. Surr. Switch & 0.8178$_{\pm 0.0316}$ & 0.9776$_{\pm 0.0194}$ & 0.9989$_{\pm 0.0000}$ & 0.9837$_{\pm 0.0028}$ & 0.9445$_{\pm 0.0849}$ & -0.97\% \\
 & w.o. LLM Acq. & 0.8250$_{\pm 0.0160}$ & 0.9101$_{\pm 0.0801}$ & 0.9955$_{\pm 0.0058}$ & 0.9705$_{\pm 0.0182}$ & 0.9253$_{\pm 0.0759}$ & -2.99\% \\
 & w.o. Harness & 0.6511$_{\pm 0.1613}$ & 0.9278$_{\pm 0.0625}$ & 0.9922$_{\pm 0.0058}$ & 0.9623$_{\pm 0.0055}$ & 0.8833$_{\pm 0.1571}$ & -7.39\% \\
\hline
\end{tabular*}
\end{table*}
\noindent\textbf{Overall effects.}
Table~\ref{tab:ablation} presents the ablation results of DASH on the chemical optimization tasks. Overall, full DASH achieves the highest average EF of $0.9538$. Removing any of the three components decreases both the average AF and EF, supporting the effectiveness of the overall design. The different changes in AF and EF further reveal that the modules contribute to distinct aspects of the optimization process.

\noindent\textbf{Component-wise effects.}
The integrated harness has the largest aggregate effect. Replacing knowledge-guided initialization with seeded random sampling and disabling memory feedback reduces AF by $22.31\%$ and EF by $7.39\%$, with degradation observed on all four tasks and the largest change on OCM. Replacing the LLM-guided acquisition controller with fixed qLogEI decreases AF by $1.91\%$ and EF by $2.99\%$, again producing consistent degradation across all tasks. Disabling surrogate switching results in smaller average reductions of $2.78\%$ in AF and $0.97\%$ in EF, although its effects vary across individual tasks. These results suggest that the harness primarily improves efficiency, the LLM-guided acquisition controller provides consistent task-level gains, and surrogate adaptation offers smaller but complementary benefits.

\subsection{LLM Sensitivity and Contamination Assessment}
\begin{figure}[t]
    \centering
    \includegraphics[width=0.95\columnwidth]{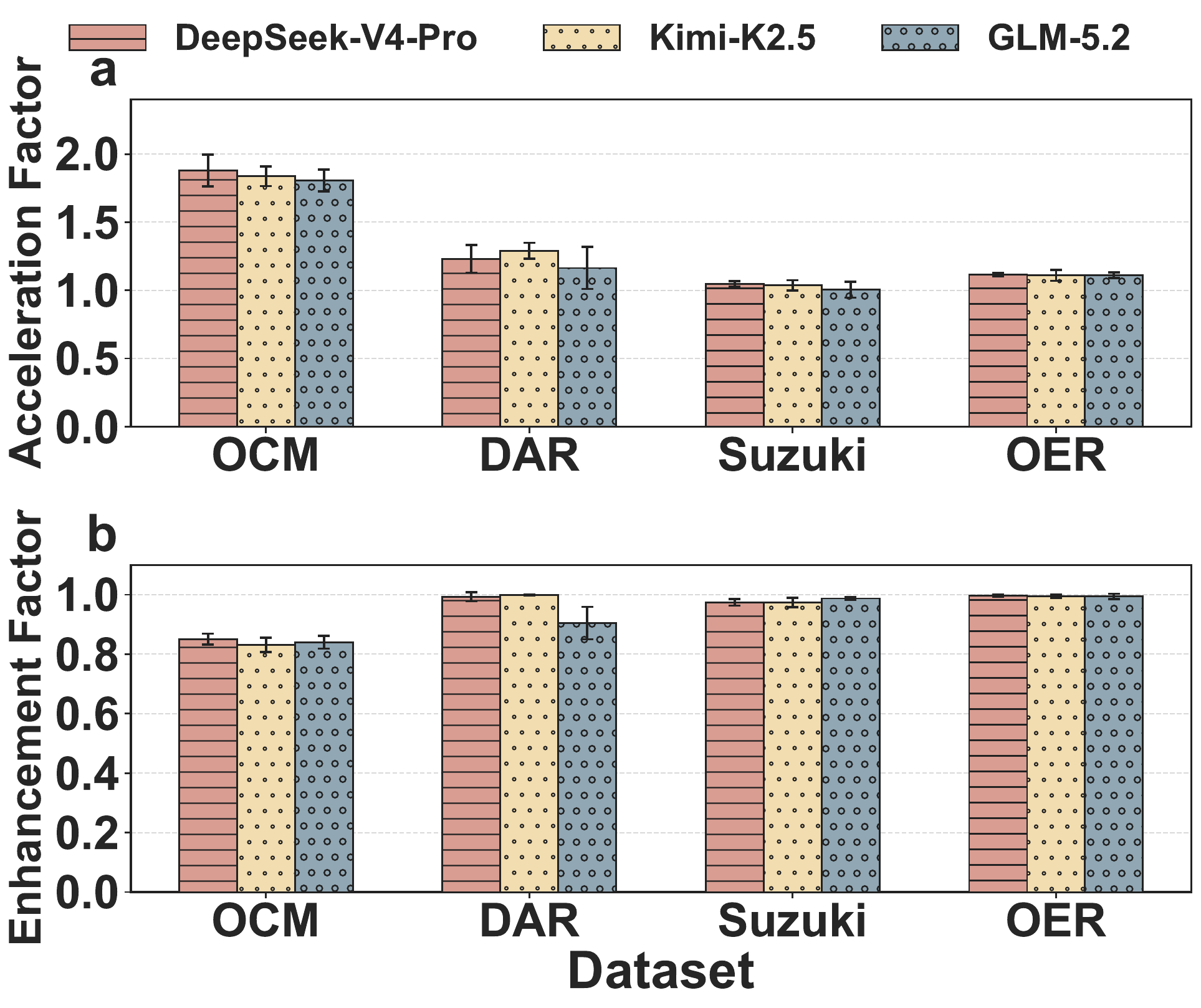}
    \caption{Effect of the LLM backbone on DASH across four chemical optimization tasks. (a) AF, where the dashed line denotes the random-search baseline; (b) EF,  where the dashed line denotes the maximum attainable value. Bars and error bars report the mean and standard deviation over five independent runs.}
    \label{fig:LLM_sensitivity}
\end{figure}

\noindent\textbf{LLM sensitivity.}
To assess the sensitivity of DASH across different LLM backbones and ensure reproducibility, we select three leading open-source LLMs---DeepSeek-V4-Pro, Kimi-K2.5, and GLM-5.2, evaluating them under identical settings on the four chemical optimization tasks. As shown in Figure~\ref{fig:LLM_sensitivity}, the three variants achieve broadly comparable AF and EF across tasks. DAR exhibits slightly greater variation, where GLM-5.2 obtains a lower EF while maintaining an AF comparable to the other models. Despite this isolated difference, no backbone consistently underperforms, demonstrating that DASH is robust to the choice of LLM backbone.
\noindent\textbf{Contamination assessment.}
\begin{table}[t]
    \centering
    \small
    \caption{Full-table memorization audit of DeepSeek-V4-Pro. DCQ values are accuracies with 95\% Wilson confidence intervals.}
    \label{tab:table_memorization_audit}
    \begin{tabular}{lccc}
        \toprule
        Dataset & DCQ Acc. (\%, 95\% CI) & Rare exact recovery\\
        \midrule
        DAR    & 12.905 [11.406, 14.568] & 0/1000 (0.000\%) \\
        OCM    & 13.865 [13.275, 14.477] & 0/243  (0.000\%) \\
        Suzuki & 22.535 [21.474, 23.632] & 0/5476 (0.000\%) \\
        OER    & 13.767 [12.366, 15.299] & 0/2085 (0.000\%) \\
        \bottomrule
    \end{tabular}
\end{table}
We audited all 22,317 source records in the four chemical datasets using two complementary protocols. The chemistry adapted Data Contamination Quiz (DCQ) asks the model to identify the original target value among chemically matched decoys \citep{golchin2025data}, whereas the Source Guided Target Slot Guessing Quiz evaluates whether the model can exactly recover a masked entry from a source table using the publication information and experimental conditions \citep{deng2024investigating,golchin2024timetravel}. We considered memorization detectable only when the DCQ Wilson lower bound exceeded the predefined 25\% threshold and at least three rare target strings were recovered exactly. LLM did not satisfy this criterion on any dataset. DAR, OCM, and OER remained below the 20\% DCQ chance level; although Suzuki reached 22.54\%, its 95\% Wilson lower bound was only 21.47\% (Table~\ref{tab:table_memorization_audit}). Moreover, none of the 8,804 rare target strings was recovered exactly. We therefore found no detectable evidence of full-table memorization or direct source-cell leakage under the evaluated protocol. 

\begin{figure}[!htbp]
    \centering
    \includegraphics[width=0.95\columnwidth]{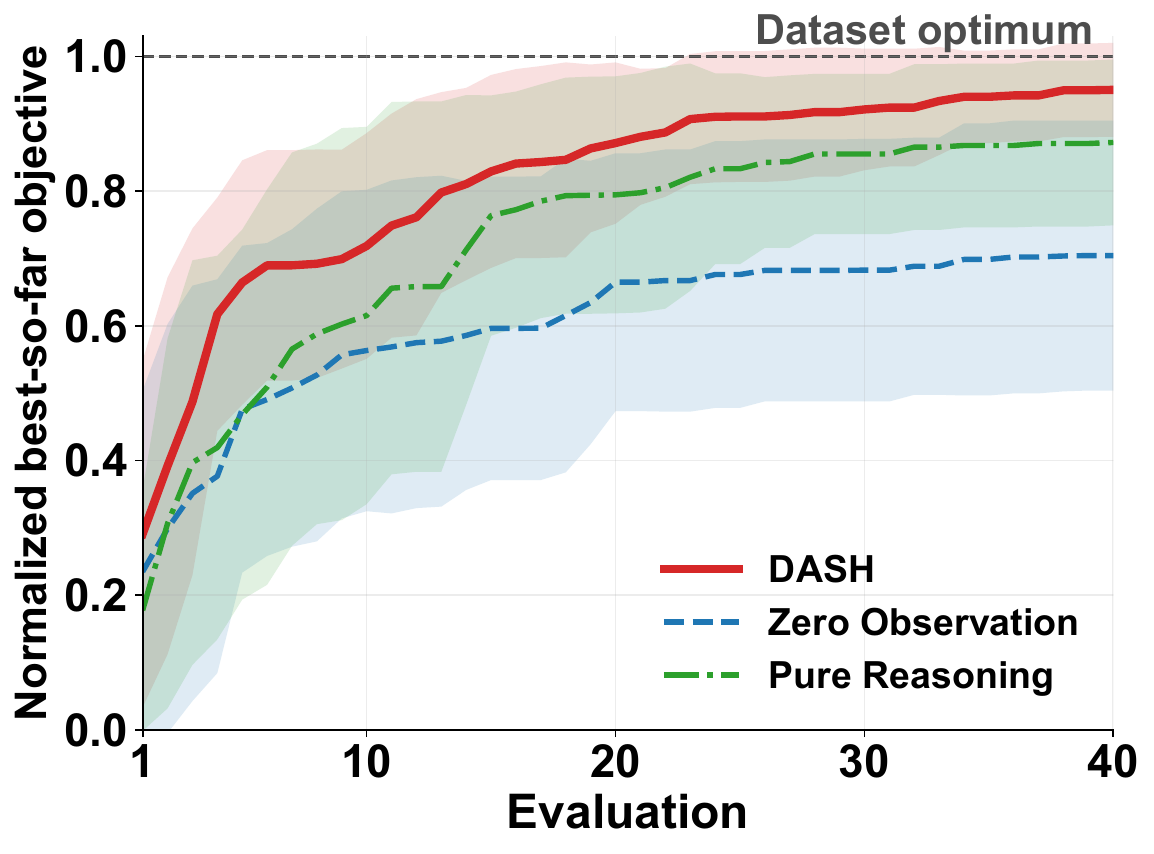}
    \caption{Behavioral check for benchmark memorization on the four chemical optimization tasks. Curves show the mean normalized best-so-far objective, with shaded regions indicating one standard deviation. Results are aggregated over four chemical optimization tasks, with five independent runs conducted for each task.}
    \label{fig:LLM_memory_test}
\end{figure}

\noindent\textbf{Behavioral memorization check.}
Figure~\ref{fig:LLM_memory_test} further examines whether the LLM can recover high-performing benchmark entries through direct reasoning alone. In Zero Observation, the LLM generates and ranks 40 candidate points in a single query without receiving any objective observations. In Pure Reasoning, each observed result is returned to the LLM before it proposes the next candidate. If benchmark targets or their rankings had been directly memorized, these protocols would be expected to recover near-optimal candidates rapidly, particularly under Zero Observation. Instead, Zero Observation remains far below the dataset optimum, while Pure Reasoning improves with sequential feedback but still underperforms DASH. Together with the full-table audit in Table~3, these results provide no detectable evidence that DASH's gains arise from direct benchmark memorization. However, they do not rule out all possible exposure of the datasets during pretraining.

\section{Conclusion}
In this paper, we study how to adapt surrogate models and acquisition strategies simultaneously during  Bayesian optimization and propose DASH, a decoupled framework that integrates LLM reasoning into this process. It comprises two complementary components: a decoupled adaptation architecture
with separate controllers for surrogate and acquisition selection, together with a set of essential harnessing mechanisms. In the adaptation architecture, the data-driven surrogate controller evaluates candidate models according to predictive reliability, uncertainty calibration, and ranking consistency, while the context-driven acquisition controller uses an LLM to allocate candidate quotas across acquisition functions and select the next evaluation from the bounded shortlist. The harnessing mechanisms combine knowledge-guided initialization with structured memory to ground the optimization process in domain knowledge and accumulated feedback. Experiments show that DASH achieves the strongest average performance across four chemical optimization benchmarks and remains highly competitive on four HPO benchmarks. Ablation studies further confirm the effectiveness of all components. Moreover, DASH maintains robust performance across different LLM backbones,
with no detectable evidence that the observed gains result from memorization of the benchmark data by the LLMs.

\bibliography{aaai2027}

\end{document}